\documentclass[letterpaper,10pt,conference]{ieeeconf}
\usepackage{times,mathptmx,amsmath,amssymb,bm,graphicx,booktabs,array,multirow,url}
\usepackage[table]{xcolor}
\usepackage{siunitx}
\let\labelindent\relax
\usepackage[inline]{enumitem}
\usepackage[hidelinks,hyperfootnotes=false]{hyperref}
\usepackage{pgfplots}
\pgfplotsset{compat=1.17}
\usepgfplotslibrary{groupplots}
\usepackage{lipsum}
\usepackage[skip=4pt,font=small]{caption}

\usepackage[hidelinks]{hyperref}
\newcommand{\SO}{\mathrm{SO}(3)}
\newcommand{\SE}{\mathrm{SE}(3)}
\newcommand{\Sim}{\mathrm{Sim}(3)}

\makeatletter
\def\bstctlcite{\@ifnextchar[{\@bstctlcite}{\@bstctlcite[@auxout]}}
\def\@bstctlcite[#1]#2{\@bsphack\@for\@citeb:=#2\do{\edef\@citeb{\expandafter\@firstofone\@citeb}%
\if@filesw\immediate\write\csname #1\endcsname{\string\citation{\@citeb}}\fi}\@esphack}
\makeatother
\title{\LARGE \bf DAVIO: Dense Monocular--Inertial SLAM with\\Feed-Forward Initialization and Pose-Conditioned Mapping}

\author{Jaafar Mahmoud, Arthur Movsesyan, Mikhail Iumanov, and Sergey Kolyubin}

\begin{document}
\maketitle
\begingroup
\renewcommand{\thefootnote}{}
\footnotetext{The authors are with the Biomechatronics and Energy-Efficient Robotics (BE2R) Lab, ITMO University, Saint Petersburg, Russia.
{\tt\small jaafar.a.mahmoud@itmo.ru}}
\endgroup
\thispagestyle{empty}\pagestyle{empty}
\bstctlcite{IEEEexample:BSTcontrol}

\begin{abstract}
A camera and an IMU are the minimal sensor setup for metric localization and dense mapping, yet classical visual--inertial filters must wait for parallax before they start and then retain only sparse landmarks. Feed-forward geometry models, in contrast, predict dense structure from a few images but provide neither metric scale nor gravity. We present DAVIO, which uses a single multi-view depth model, Depth Anything~3, for both start-up and mapping. At start-up, a five-image window and preintegrated IMU measurements form a feature-free linear system. Its robust, conditioning-checked solution bootstraps a VIO filter through buffered replay. During tracking, the filter's metric poses condition the depth model. Residual scale is corrected only along viewing rays, which preserves the metric camera baselines, and a gravity-preserving submap graph with drift-gated revisits refines the map. On EuRoC, DAVIO starts markedly earlier, reduces the localization error, and maps more accurately than SOTA feed-forward mappers given identical poses. On building-scale ORI sequences, DAVIO is on bar or better than SOTA mappers on the same odometry, and degrades far less when GT poses are replaced by real odometry. We release the code of DAVIO, a real-time dense metric SLAM system, to the community~\footnote{https://be2rlab.github.io/DAVIO/}.
\end{abstract}

\section{Introduction}
Efficient $3D$ perception in autonomous systems heavily relies on combining visual and inertial sensing, as a single camera paired with an IMU provides sufficient information for metric scale recovery and dense reconstruction. However, inertial measurements can resolve metric scale only after sufficient motion separates specific force from gravity~\cite{martinelli2014closed}. Once initialized, classical visual–inertial odometry (VIO)~\cite{mourikis2007msckf,qin2018vinsmono,geneva2020openvins,campos2021orbslam3,peng2024sqrtvins} algorithms maintain tracked sparse landmarks. While sufficient for trajectory estimation, these representations are far too sparse for downstream robotic applications such as autonomous navigation.


\begin{figure}[t]
\centering
\includegraphics[width=\columnwidth]{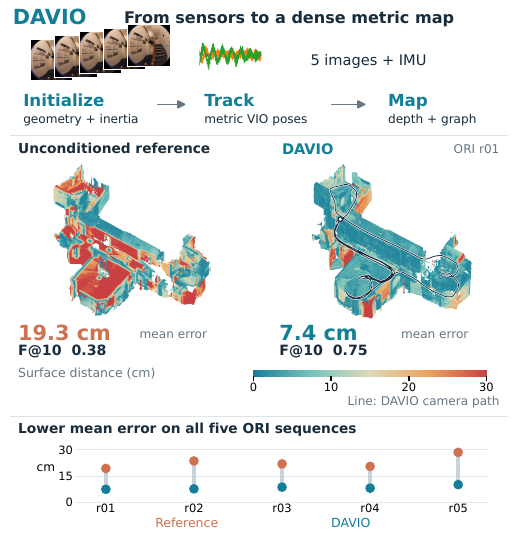}
\caption{\textbf{From a monocular camera and an IMU to a dense metric map.} DAVIO
initializes from images and IMU measurements, then conditions dense mapping on metric VIO poses through incremental submap fusion.}
\label{fig:teaser}
\end{figure}

Feed-forward geometry models predict dense point maps or depth directly from co-visible images in a single forward
pass~\cite{wang2024dust3r,leroy2024mast3r,wang2025vggt,wang2025pi3,keetha2026mapanything,lin2025da3}.
Trained as foundation models on vast and diverse datasets, they offer strong generalizability and visual robustness. However, their predictions are strictly uncalibrated -- lacking metric scale, gravity orientation, sensor velocity, and IMU biases. The two modalities are thus highly complementary: inertial measurements supply the physical quantities missing in learned geometry, while feed-forward models predict dense structure from the initial frames, well before sufficient parallax accumulates for classical initialization~\cite{zhang2026ffvioinit}.


To bridge these complementary strengths while addressing their individual limitations, we propose \textbf{DAVIO}, a monocular-inertial system that fuses feed-forward visual geometry with filter-based VIO across two distinct operational phases. 

During initialization, DAVIO constructs a feature-free state proposal from a five-image window and preintegrated IMU measurements following~\cite{zhang2026ffvioinit} (Sec.~\ref{sec:ffinit}). This proposal bootstraps an unmodified OpenVINS filter~\cite{geneva2020openvins} via a buffered-replay mechanism (Sec.~\ref{sec:refinement}). Sensor data arriving during initialization are queued and replayed faster than real time, and the filter publishes only once it has caught up with the live stream, so no measurements are dropped.

During tracking, the interaction reverses: the filter’s metric camera poses condition the geometric foundation model (DA3~\cite{lin2025da3}), yielding overlapping depth submaps (Sec.~\ref{sec:da3-inference}). Residual scale corrections are applied only along viewing rays at depth back-projection, thereby preserving the filter’s metric camera baselines (Sec.~\ref{sec:framescale}). Additionally, a gravity-preserving submap graph jointly refines depth scales and map-frame poses, admitting revisit constraints to the pose problem only after geometric verification and a drift test  (Sec.~\ref{sec:graph}). The dense map is thus available within seconds of start-up, while the filter state is never modified; the map-frame correction lowers the final trajectory error.

Our main contributions are summarized as follows:
\begin{enumerate}
    \item \textbf{Dual-purpose framework architecture:} A real-time unified monocular--inertial system that uses one feed-forward geometry model for both initialization and dense mapping.
    \item \textbf{Robust point--inertial initializer:} An extension of the feature-free formulation of~\cite{zhang2026ffvioinit}: a stratified, scale-normalized LMedS search with conditioning and physical-plausibility gates, designed to reject unreliable windows rather than pass a wrong scale to the VIO filter.
    \item \textbf{Pose-conditioned mapping formulation:} A depth-scaling strategy that applies residual scale corrections only in camera space -- preserving intra-window metric baselines -- coupled with a submap graph in which relative-scale constraints and gravity-preserving pose updates pass through separate gates.
\end{enumerate}

\section{Related Work}\label{sec:related}
\textbf{VIO and initialization.} Filter-based VIO propagates inertial measurements
against tracked features~\cite{mourikis2007msckf,geneva2020openvins}, whereas
optimization-based VIO solves a sliding-window
problem~\cite{qin2018vinsmono,campos2021orbslam3}. Initializers align a visual-only map
with inertial measurements~\cite{campos2020inertialonly}, decouple rotation from
translation so that scale and gravity enter a small linear
problem~\cite{he2023drt,peng2024sqrtvins}, or refine a closed-form
solution~\cite{cerezo2025closedform}. Learned depth priors have entered this
problem~\cite{zhou2022learneddepthinit}, and Zhang et al.~\cite{zhang2026ffvioinit}
replace tracked landmarks with feed-forward point clouds, removing feature tracking
from start-up. DAVIO adopts this formulation and extends it with key components required for continuous operation:
frame-balanced robust estimation, conditioning checks, and a buffered filter handoff,
while online extrinsic and temporal calibration remain with the
filter~\cite{geneva2020openvins,yang2023selfcal}.

\textbf{Learned dense geometry.} DUSt3R~\cite{wang2024dust3r} and MASt3R~\cite{leroy2024mast3r} regress pairwise point maps in a
shared frame; VGGT~\cite{wang2025vggt} and $\pi^3$~\cite{wang2025pi3} generalise this formulation to multi-view input; MapAnything~\cite{keetha2026mapanything} and DA3~\cite{lin2025da3} also accept camera
intrinsics and poses. SLAM systems built on
these predictors jointly optimize pose and geometry~\cite{teed2021droid}, incrementally
align feed-forward
submaps~\cite{murai2025mast3rslam,maggio2025vggtslam,maggio2026vggtslam2,lee2026unisim},
or retain a lightweight feature tracker~\cite{hu2026ec3rslam}. Being monocular, they
recover scale per submap, naturally in $\Sim$. In DAVIO the translations inside each
window are already metric, so it keeps them and applies residual scale only where the
prediction is uncertain, at depth back-projection.

\textbf{Learned mapping anchored to inertial motion.} SimpleMapping couples VIO with
multi-view stereo~\cite{xin2023simplemapping}, Kimera adds a dense mesh layer to a
visual--inertial pipeline~\cite{rosinol2020kimera}, MASt3R-Fusion couples learned
alignment with IMU and GNSS factors~\cite{zhou2026mast3rfusion}, and VIDAR anchors
learned mapping to VIO~\cite{salih2026vidar}. Most closely related to DAVIO, ScaRF-SLAM conditions
DA3 on classical tracking poses and optimizes frame and submap scales with those poses
fixed~\cite{zhang2026scarfslam}. 
DAVIO similarly leaves the online VIO filter state untouched.
However, it differs in two key aspects: first, DAVIO bootstraps the filter using predictions
from the very same feed-forward model; second, it incorporates a gravity-preserving submap graph
that refines map-frame poses during verified revisits, admitting loop closures only when the
accumulated odometry drift exceeds the constraint's uncertainty.

\section{System Overview}\label{sec:overview}
\begin{figure*}[t]
\centering
\includegraphics[width=0.9\textwidth]{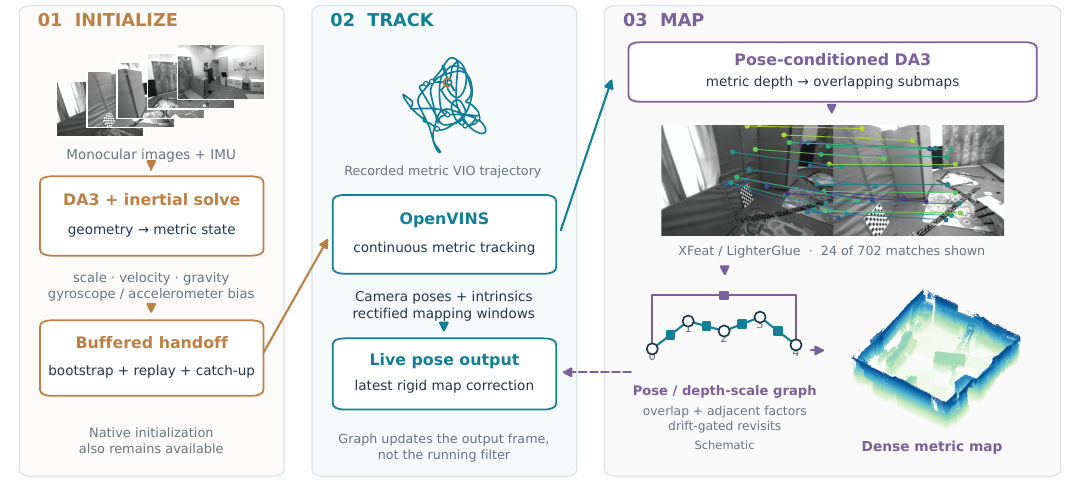}
\caption{\textbf{DAVIO in three stages.} (1)~A feed-forward visual--inertial solve on a five-image window proposes a metric state that bootstraps a fresh OpenVINS~\cite{geneva2020openvins} filter through buffered replay and catch-up checks. (2)~The filter tracks continuously and supplies metric camera poses and intrinsics.
(3)~Pose-conditioned DA3 windows form overlapping submaps; in-window matches refine per-frame depth scales, and shared-image ratios and verified correspondences constrain a gravity-preserving submap graph whose revisits move poses only past a drift gate.}
\label{fig:system}
\end{figure*}
\textbf{Notation.} Bold lowercase symbols denote vectors, and $\log$ is the natural logarithm. For frames $A$ and $B$, $T_{AB}\in\SE$ maps coordinates expressed in $B$ to coordinates expressed in $A$, $\mathbf x_A=T_{AB}\,\mathbf x_B$. It consists of the rotation $R_{AB}\in\SO$ and the position $\mathbf p_{AB}\in\mathbb R^3$ of the origin of $B$ expressed in $A$, so that $T_{BA}=T_{AB}^{-1}$ and $R_{BA}=R_{AB}^\top$. The symbols $I$, $C$, $O$, and $M$ denote the IMU, the camera, the odometry frame of the VIO filter, and the optimized map frame; $I_i$ and $C_i$ denote the IMU and the camera at image $i$, with $i=0$ the reference image. The camera--IMU extrinsics are the rotation $R_{IC}$ and the lever arm $\mathbf p_{IC}$, i.e., the camera origin expressed in the IMU frame, so the camera pose reported by the filter is $T_{OC}(t)=T_{OI}(t)\,T_{IC}$.

\textbf{Architecture.} A tracking process runs filter-based visual--inertial odometry (OpenVINS~\cite{geneva2020openvins}), while a single worker hosts the GPU model (Fig.~\ref{fig:system}). At start-up, five images spaced 0.2\,s apart form a geometry window; if its estimate is accepted, it initializes a fresh filter. The OpenVINS initializer remains armed, and whichever filter becomes ready first publishes. During tracking, the worker forms five-frame windows at 0.5\,s keyframe spacing, with two images shared between consecutive windows, and commits one map version per window. Each commit defines a rigid correction $T_{MO}$ that is applied to subsequently published poses, whereas velocities, biases, and calibration parameters remain in the filter.

\section{Feed-Forward Initialization}\label{sec:ffinit}

\subsection{Point--inertial linear system}
For the start-up window, DA3~\cite{lin2025da3} predicts per-image depth, confidence, intrinsics, and relative camera poses. Up to 100 points per image whose confidence exceeds the 25th percentile are sampled over a $4\times4$ grid. Let $\mathbf z_{ik}\in\mathbb R^2$ be the normalized coordinates of pixel $k$ in image $i$, $\tilde{\mathbf z}_{ik}=[\mathbf z_{ik}^\top,1]^\top$, and $\bar d_{ik}$ its predicted up-to-scale depth. Using DA3's relative pose $(R_{C_0C_i},\bar{\mathbf p}_{C_0C_i})$, the pixel back-projects to the up-to-scale point
\begin{equation}
\bar{\mathbf p}_{ik}=R_{C_0C_i}\,\bar d_{ik}\,\tilde{\mathbf z}_{ik}+\bar{\mathbf p}_{C_0C_i}
\end{equation}
expressed in $C_0$. Following~\cite{calibmorkis}, define
\begin{equation}
H_{ik}=\begin{bmatrix}1&0&-z_{ik,x}\\0&1&-z_{ik,y}\end{bmatrix},
\end{equation}
so that $H_{ik}\mathbf y=\mathbf 0$ for every point $\mathbf y$, expressed in $C_i$, that lies on the viewing ray of $\mathbf z_{ik}$.

IMU preintegration~\cite{forster2017preintegration} at the current bias estimate yields the relative rotation $R_{I_0I_i}$ and the gravity-free displacement $\Delta\mathbf p_i$, i.e., the double integral of the rotated specific force, expressed in $I_0$ over the interval $\Delta t_i$. Let $\mathbf v_0$ and $\mathbf g$ denote the velocity of $I_0$ and the gravity vector, both expressed in $I_0$. As in OpenVINS, $\mathbf g$ points upward, so the gravitational acceleration is $-\mathbf g$, with $\|\mathbf g\|=g_0$; at rest, the measured specific force expressed in $I_0$ is $+\mathbf g$. The position of $I_i$ expressed in $I_0$ is then 
\begin{equation}
\mathbf p_{I_0I_i}=\Delta t_i\,\mathbf v_0-\tfrac12\Delta t_i^2\,\mathbf g+\Delta\mathbf p_i .
\label{eq:imupos}
\end{equation}
For a metric scale $s>0$, chaining $C_0\!\to\!I_0\!\to\!I_i\!\to\!C_i$ expresses the point in camera $i$ as 
\begin{equation}
\mathbf y_{ik}=R_{CI}\Bigl[R_{I_iI_0}\bigl(s\,R_{IC}\,\bar{\mathbf p}_{ik}+\mathbf p_{IC}
-\mathbf p_{I_0I_i}\bigr)-\mathbf p_{IC}\Bigr].
\label{eq:initpoint}
\end{equation}
Substituting~\eqref{eq:imupos} shows that the ray constraint $H_{ik}\mathbf y_{ik}=\mathbf 0$ is linear in $\mathbf x=[s,\mathbf v_0^\top,\mathbf g^\top]^\top\in\mathbb R^7$, i.e., $A_{ik}\mathbf x=\mathbf b_{ik}$ with $B_{ik}=H_{ik}R_{CI}R_{I_iI_0}$ and 
\begin{align}
A_{ik}&=\Bigl[\,B_{ik}R_{IC}\bar{\mathbf p}_{ik},\;\; -\Delta t_i\,B_{ik},\;\;
\tfrac12\Delta t_i^2\,B_{ik}\Bigr],\nonumber\\
\mathbf b_{ik}&=B_{ik}\,\Delta\mathbf p_i-H_{ik}R_{CI}\bigl(R_{I_iI_0}-I_3\bigr)\mathbf p_{IC}.
\label{eq:linear}
\end{align}
Since $H_{ik}\mathbf y_{ik}=y_{ik,z}\bigl(\pi(\mathbf y_{ik})-\mathbf z_{ik}\bigr)$ and $y_{ik,z}\approx s\,\bar d_{ik}$, each row pair is divided by $\bar d_{ik}$ and weighted by the square root of the point's confidence. The resulting algebraic residual therefore approximates $s$ times the normalized reprojection error. The lever arm $\mathbf p_{IC}$, which appears explicitly in~\eqref{eq:linear}, is fixed at its supplied value (zero when withheld; Sec.~\ref{sec:modes}). The weighted least-squares solution is followed by two refinements of the gravity direction on the tangent plane of the sphere $\|\mathbf g\|=g_0$.

\subsection{Conditioning and robust estimation}
Although DA3~\cite{lin2025da3} provides hundreds of points per image, the system
in~\eqref{eq:linear} is far less overdetermined than it appears. If DA3's relative
rotations agree with the preintegrated ones, $R_{C_0C_i}=R_{CI}R_{I_0I_i}R_{IC}$, the depth component of each point lies on its own viewing ray and is annihilated by $H_{ik}$.
Equation~\eqref{eq:initpoint} then reduces to $H_{ik}\mathbf y_{ik}=B_{ik}\mathbf w_i$, with \begin{equation}
\mathbf w_i=s\,R_{IC}\,\bar{\mathbf p}_{C_0C_i}-\Delta t_i\,\mathbf v_0
+\tfrac12\Delta t_i^2\,\mathbf g-\Delta\mathbf p_i-\bigl(R_{I_0I_i}-I_3\bigr)\mathbf p_{IC}.
\label{eq:frameblock}
\end{equation}
Thus all points of image $i$ test a single statement: the scaled DA3 camera center must match the IMU-integrated position. Because $H_{ik}$ has rank two, two points with distinct rays already determine $\mathbf w_i$, which is affine in $\mathbf x$ through a $3\times7$ block. Each image therefore contributes at most three independent equations to the seven unknowns, however many points it has; since $\mathbf w_0=\mathbf 0$, at least four images are required, and the five-image window adds one image of redundancy. Dense points and confidence weights improve robustness to outliers and noise but do not raise this rank. Rotation disagreement between DA3 and the IMU can raise the numerical rank, but the added rows reflect rotation error, not scale information.

\textbf{Degenerate motion.}
Four images are necessary but not sufficient. If the up-to-scale trajectory is at most quadratic in time, $\bar{\mathbf p}_{C_0C_i}=\Delta t_i\,\mathbf a+\Delta t_i^2\,\mathbf c$, then under the same rotation assumption
\begin{equation}
A_{ik}\bigl[\,1,\;(R_{IC}\mathbf a)^\top,\;(-2R_{IC}\mathbf c)^\top\bigr]^\top=\mathbf 0
\quad\forall\,(i,k),
\label{eq:degenerate}
\end{equation}
so a change in $s$ is absorbed by $\mathbf v_0$ and $\mathbf g$. Constant-velocity motion ($\mathbf c=\mathbf 0$), which is common at start-up, is the typical instance. We therefore accept the linear solution only if
(i) at least four images are available,
(ii) the column-equilibrated system in $\mathbf x$ has a singular-value ratio of at least $10^{-3}$, and
(iii) the baseline-to-depth ratio is at least 0.02.
These are numerical safeguards, not an observability proof, because the constraint
$\|\mathbf g\|=g_0$ may restore observability in some rejected cases. Conservative
rejection is inexpensive, since the OpenVINS initializer remains armed.

\textbf{Robust estimation.}
DA3 depth is unreliable at object boundaries, on reflective or textureless surfaces, and on moving objects. Since the outlier fraction is unknown, we use a least-median-of-squares (LMedS) search over 100 hypotheses. Each hypothesis is fitted to two points from every non-reference image. Two points already saturate an image's rank-three contribution, so this stratified sample is well posed and prevents any single image from dominating. Because the algebraic residual $e_k$ scales with $s$, hypotheses with $s>0$ are scored by $\operatorname{med}_k e_k/s$, which removes the bias toward small scale. A hypothesis is admissible only if at least three images keep a majority of their points within $\max(2.5\,\hat\sigma,\,0.01)$, where $\hat\sigma=1.4826\,\operatorname{med}_k e_k$.

\textbf{Acceptance.}
The best hypothesis is refitted on its inliers. It is accepted only if it is well
supported, with an inlier fraction of at least 0.4 and a median residual of at most 0.05. It must also be physically plausible, with $\|\mathbf v_0\|\le5$\,m/s and $\mathbf g$ within $30^\circ$ of the mean specific-force direction. These tests convert silent failures into explicit rejections, while the OpenVINS initializer remains armed. The gyroscope bias is initialized by aligning DA3's relative rotations with $R_{I_0I_i}$, which gives the preintegration a better linearization point than zero bias.

\subsection{Refinement and filter handoff}\label{sec:refinement}
The accepted estimate is refined over $\boldsymbol\theta=[\eta,\mathbf v_0^\top,\delta\boldsymbol\gamma^\top, \delta\mathbf b_g^\top,\mathbf b_a^\top]^\top$. The softplus parameterization $s=\epsilon+\log(1+e^{\eta})$, with $\epsilon=10^{-5}$, keeps the scale strictly positive. The gravity vector is parameterized by two tangent coordinates about the linear-stage direction $\hat{\mathbf g}$, 
\begin{equation}
\mathbf g(\delta\boldsymbol\gamma)=g_0\,\mathrm{Exp}\bigl(N\delta\boldsymbol\gamma\bigr)\,
\hat{\mathbf g},\qquad N\in\mathbb R^{3\times2},\;\; N^\top\hat{\mathbf g}=\mathbf 0 .
\end{equation}
Here $\delta\mathbf b_g$ is the gyroscope-bias change relative to the preintegration estimate, and $\mathbf b_a$ is the accelerometer bias, with
$\delta\mathbf b_a=\mathbf b_a-\bar{\mathbf b}_a$. The preintegrated terms are corrected to first order~\cite{forster2017preintegration}:
\begin{align}
R_{I_iI_0}(\boldsymbol\theta)&\approx\mathrm{Exp}\bigl(-J^{R}_{i}\,\delta\mathbf b_g\bigr)\,
\bar R_{I_iI_0},\nonumber\\
\Delta\mathbf p_i(\boldsymbol\theta)&\approx\Delta\bar{\mathbf p}_i
+J^{p}_{g,i}\,\delta\mathbf b_g+J^{p}_{a,i}\,\delta\mathbf b_a ,
\end{align}
and $\mathbf y_{ik}(\boldsymbol\theta)$ follows from~\eqref{eq:imupos}
and~\eqref{eq:initpoint}. Let $\pi(\mathbf y)=[y_x,y_y]^\top/\max(y_z,10^{-6})$ be the normalized projection, whose clamp only guards the division (a point behind the camera receives a large residual of bounded influence), and $s_{\rm lin}$ the linear-stage scale. The stacked residual over the inlier set $\mathcal K$ is:
\begin{equation}
\mathbf r(\boldsymbol\theta)=\begin{bmatrix}
\bigl\{\sigma_u^{-1}\bigl(\pi(\mathbf y_{ik}(\boldsymbol\theta))-\mathbf z_{ik}\bigr)
\bigr\}_{(i,k)\in\mathcal K}\\[2pt]
\sigma_g^{-1}\,\delta\mathbf b_g\\[2pt]
\sigma_a^{-1}\,\mathbf b_a\\[2pt]
\sigma_s^{-1}\log(s/s_{\rm lin})
\end{bmatrix},
\label{eq:refine}
\end{equation}
and is minimized under a component-wise Huber loss with unit threshold, using
$\sigma_u=0.007$, $\sigma_g=0.05$\,rad/s, $\sigma_a=0.02$\,m/s$^2$, and $\sigma_s=0.5$.
The priors are essential: within a sub-second window, the accelerometer bias is only weakly separable from scale and gravity, and the scale prior keeps the refined scale close to the accepted linear estimate.

The refined state is propagated to the last image and expressed in a gravity-aligned world frame. It is then written once, with block-wise covariance floors, into a fresh OpenVINS instance through the unmodified initialization entry point. The new instance replays up to 8\,s of buffered measurements and is selected once it has produced five consecutive states within 0.15\,s of the live stream. A candidate that fails to do so within 3\,s is abandoned, and at most five candidates are attempted.

\subsection{Calibration modes}\label{sec:modes}
In the primary mode, intrinsics, distortion, extrinsics, and time offset are taken from the rig calibration. A second mode withholds the extrinsics and time offset. The rotation $R_{IC}$ and the gyroscope bias $\mathbf b_g$ are then estimated from several early windows by aligning preintegrated and DA3 relative rotations, i.e., by minimizing the hand--eye residuals
\begin{equation}
\mathbf r^{\rm he}_{ii'}=\mathrm{Log}\bigl(R_{I_{i'}I_i}(\mathbf b_g)\,R_{IC}\,
R_{C_iC_{i'}}\,R_{CI}\bigr),
\label{eq:handeye}
\end{equation}
which vanish when $R_{I_{i'}I_i}=R_{IC}R_{C_{i'}C_i}R_{CI}$. The estimate is validated on a later window. The lever arm $\mathbf p_{IC}$ and the time offset are initialized to zero, and OpenVINS~\cite{geneva2020openvins} refines all three quantities online from priors.

\section{Pose-Conditioned Dense Mapping}\label{sec:mapping}
\subsection{Conditioned submaps and depth-only scale}\label{sec:da3-inference}
For each window, the worker passes DA3 the images, the rectified intrinsics $K$, and the filter's metric poses $T_{C_iO}$. A similarity fit between the returned and the VIO camera centers gives an initial depth correction $\hat s_j$. The window is accepted if its metric baseline exceeds 5\,cm, the fit RMSE is at most 15\,cm, and $|\log\hat s_j|\le0.3$. Since DA3's default interface returns the input poses, this fit serves as a consistency check rather than as independent evidence of depth accuracy.

Let $S_j$ denote the camera frame of the center image of submap $j$. Each submap carries a pose $Q_j=T_{MS_j}\in\SE$ and a residual log-depth scale $\lambda_j$, while $L_{ji}=T_{S_jC_i}$ is the fixed metric VIO pose of image $i$ relative to the center. With the corrected depth map $d_{ji}=\hat s_j\,s_{ji}\,d^{\rm DA3}_{ji}$, where $s_{ji}$ is a per-frame scale (Sec.~\ref{sec:framescale}), a homogeneous pixel $\mathbf u$ is placed in the map at 
\begin{equation}
\mathbf p_M=T_{MS_j}\,T_{S_jC_i}\bigl(e^{\lambda_j}\,d_{ji}(\mathbf u)\,K^{-1}\mathbf u\bigr).
\label{eq:placement}
\end{equation}
Scale acts on depth only, so the metric translations of $T_{S_jC_i}$ are never rescaled. A similarity applied to the whole window would instead scale the VIO baselines and displace every non-center frame.

\subsection{Frame scales and shared-image constraints}\label{sec:framescale}
XFeat/LighterGlue matches~\cite{potje2024xfeat,lindenberger2023lightglue} link images one or two positions apart within a window. For match $k$ between images $i$ and $i'$, with back-projected points $\mathbf p_{ik}=\hat s_j d^{\rm DA3}_{ji}(\mathbf u_{ik})K^{-1} \mathbf u_{ik}$ in $C_i$ (and $\mathbf p_{i'k}$ analogously), the residual 
\begin{equation}
\mathbf r_{ii'k}=T_{S_jC_i}\bigl(s_{ji}\,\mathbf p_{ik}\bigr)
-T_{S_jC_{i'}}\bigl(s_{ji'}\,\mathbf p_{i'k}\bigr)
\label{eq:framescale}
\end{equation}
is affine in the per-frame scales, as in the frame-level refinement of ScaRF-SLAM~\cite{zhang2026scarfslam}. We run eight rounds of depth-normalized, outlier-reweighted least squares with a weak unit prior. After each round, the scales are projected onto $\sum_i\log s_{ji}=0$, i.e., normalized to unit geometric mean. The per-frame scales thus redistribute only relative corrections within the window, while the overall scale of the window is estimated by $\lambda_j$ in the submap graph (Sec.~\ref{sec:graph}).

Consecutive submaps $a$ and $b$ share the images $\mathcal I_{ab}$, of which there are two unless the queue dropped a window. For each shared image $i$, the median log-depth ratio over aligned valid pixels is
$m_i=\operatorname{med}_{\mathbf u}\log\bigl(d_{bi}(\mathbf u)/d_{ai}(\mathbf u)\bigr)$.
Consistent corrected depths, $e^{\lambda_a}d_{ai}=e^{\lambda_b}d_{bi}$, imply
$\lambda_b-\lambda_a=-m_i$, which gives the relative-scale residual~\cite{lee2026unisim}
\begin{equation}
r^s_{ab}=\lambda_b-\lambda_a-\kappa_{ab},\qquad
\kappa_{ab}=-\frac{1}{|\mathcal I_{ab}|}\sum_{i\in\mathcal I_{ab}}m_i,
\label{eq:shared}
\end{equation}
with $\sigma_{ab}=\max\bigl(0.03,\tfrac12[\max_i m_i-\min_i m_i]\bigr)$. If no shared image remains, $\kappa_{ab}$ is obtained from a projected-overlap depth ratio instead.

\subsection{Submap graph and revisit admission}\label{sec:graph}
Correspondences between center images connect adjacent submaps, as well as revisits retrieved within 8\,m and at least 5\,s apart. A revisit candidate must pass PnP RANSAC~\cite{lepetit2009epnp} ($\ge$50 matches, $\ge$30 inliers, inlier fraction $\ge$0.25, 2\,px threshold), followed by positive-depth, image-coverage, depth-agreement (median relative error $\le$0.2), and $3\sigma$ odometry-innovation tests. Up to 80 verified correspondences then form a point group. PnP serves only for verification; its pose is not added as a second factor, since it derives from the same matches.
Let $Z_{ab}=T_{S_aS_b}$ be the VIO pose of center $b$ relative to center $a$, with translation $\mathbf t_{ab}$, and let $\mathbf p_{ak}$ and $\mathbf p_{bk}$ be the corresponding points in $S_a$ and $S_b$. Since $Q_a^{-1}Q_b=T_{S_aM}T_{MS_b}$ predicts $Z_{ab}$, the point and odometry residuals are:
\begin{align}
\mathbf r^p_{abk}&=T_{MS_a}\bigl(e^{\lambda_a}\mathbf p_{ak}\bigr)
-T_{MS_b}\bigl(e^{\lambda_b}\mathbf p_{bk}\bigr),\nonumber\\
\mathbf r^o_{ab}&=\mathrm{Log}\bigl(Z_{ab}^{-1}\,T_{S_aM}\,T_{MS_b}\bigr).
\label{eq:graphres}
\end{align}
The graph minimizes
\begin{align}
E={}&\sum_{\mathcal E_o}\rho\bigl(\|W_o\mathbf r^o_{ab}\|\bigr)
+\sum_{\mathcal E_p}\frac{1}{N_{ab}}\sum_{k=1}^{N_{ab}}
\rho\bigl(\|W_{abk}\mathbf r^p_{abk}\|\bigr)\nonumber\\
&+\sum_{j>0}\rho\Bigl(\frac{|\lambda_j-\lambda_0|}{0.10}\Bigr)
+\sum_{\mathcal E_s}\rho\Bigl(\frac{|r^s_{ab}|}{\sigma_{ab}}\Bigr),
\label{eq:objective}
\end{align}
where $\rho$ is the Huber loss with threshold 2, i.e., $\rho(r)=r^2/2$ for $r\le2$ and $2r-2$ otherwise.

The weights encode the following models. $W_o$ whitens with a prescribed drift model, with standard deviations $\max(2\,\mathrm{cm},\,0.005\|\mathbf t_{ab}\|)$ in translation and $\max(0.5^\circ,\,0.5^\circ\mathrm{m}^{-1}\|\mathbf t_{ab}\|)$ in rotation. $W_{abk}$ whitens with a ray-aligned covariance, with 10\,\% of depth along the ray and 1\,\% across it, both floored at 10\,cm. This covariance is frozen within a solve, while the robust weights are updated. The set $\mathcal E_s$ contains the shared-image residuals~\eqref{eq:shared} and the median depth ratios of verified revisits ($\sigma_{ab}=0.03$).

The first node fixes the gauge. It starts with $\lambda_0=0$, and after eviction the first retained node is fixed at its current state. Every other node optimizes translation, yaw about gravity, and $\lambda_j$, so the roll and pitch observed by the filter are preserved. Whenever a scale or point constraint is added, up to eight Levenberg--Marquardt iterations are run. Scale and pose constraints are gated separately: every verified revisit adds its median depth ratio to $\mathcal E_s$, but its point group enters $\mathcal E_p$, and can move the poses, only if the odometry drift between the two submaps, predicted by the drift model of $W_o$, exceeds the uncertainty of the revisit constraint. Adjacent point groups are always included.

\subsection{Fusion and live trajectory}
A depth is retained only if another frame of the same window re-observes it within
10\,\% relative depth. Retained depths are archived in submap coordinates and fused into 4\,cm voxels over at most 24 resident submaps. The final export reprojects all archives with the latest graph states.

Each commit defines the odometry-to-map correction from its newest node $j$,
\begin{equation}
T_{MO}=T_{MS_j}\,T_{OS_j}^{-1}=T_{MS_j}\,T_{S_jO},
\label{eq:correction}
\end{equation}
and live poses are published as $T^{\rm live}_{MI}(t)=T_{MO}\,T_{OI}(t)$. If the
correction changes by more than 5\,cm or $2^\circ$, it is interpolated over one sensor second, linearly in translation and along the geodesic on $\SO$ in rotation. Each published pose records its correction version and interpolation fraction, and the final export applies the completed corrections to all past poses.

\section{Experimental Setup}\label{sec:setup}
\textbf{Research questions.} We ask whether DAVIO (Q1)~provides a metric state earlier than the OpenVINS initializer, (Q2)~improves the retrospective trajectory, and at what cost to the causal one, and (Q3)~improves dense reconstruction under identical inputs. \emph{Native} is the trajectory of OpenVINS~\cite{geneva2020openvins} running alone, with its own initializer and mapping disabled (median of three runs). For a DAVIO run, \emph{causal} is the published map-frame trajectory, in which each pose uses the latest map correction available at publication time, and \emph{final} the trajectory re-exported with all corrections after the last graph update.

\textbf{Data and platform.} EuRoC~\cite{burri2016euroc} provides eleven sequences in a Machine Hall (MH) and two Vicon rooms, with OpenVINS reference trajectories, including the corrected V1\_01 reference~\cite{burri2016euroc}, and Leica scans of the six Vicon-room sequences. ORI~\cite{zhang2026scarfslam} provides five building-scale sequences (R01--R05) with survey-grade LiDAR references. DAVIO runs on an i7-12700H laptop with an RTX~3060 Laptop GPU (6\,GB) and DA3-Base at 504\,px, paced at sensor rate on EuRoC and on ORI (with resizing and rectification, as DA3~\cite{lin2025da3} operates on undistorted images).

\textbf{Comparison categories.} Tables label three categories. \emph{Matched-pose} comparisons isolate the mapping layer: ScaRF-SLAM's released mapper~\cite{zhang2026scarfslam} runs offline on DAVIO's OpenVINS poses with the same DA3-Base checkpoint. \emph{End-to-end} comparisons use each system's own tracker: ORB-SLAM3~\cite{campos2021orbslam3} (monocular--inertial, stock EuRoC configuration, three runs), MASt3R-Fusion~\cite{zhou2026mast3rfusion} (EuRoC monocular--inertial configuration, every second image), and VGGT-SLAM2~\cite{maggio2026vggtslam2} (default parameters, keyframes, and one global $\Sim$ fit to the reference, which favors it); the latter two run offline from released code with their own confidence filters, on MH\_01, MH\_03, MH\_05, and ORI.


\textbf{Metrics.} Start-up follows~\cite{zhang2026ffvioinit}: 132 ten-second windows per initializer (one start every 10\,s; 67 in the Machine Hall, 65 in the Vicon rooms). A window succeeds if a trajectory published within it has an ATE below 0.5\,m; $T_f$ is the sensor time of the first published state (10\,s if none), and the quality-penalized time $T_q$ also charges 10\,s to every failure. ATE is the translational RMSE after one rigid alignment without scale~\cite{zhang2018tutorial}. Coverage is the fraction of the sequence covered by published poses, crediting at most 0.1\,s per pose (for baselines, the fraction of frames with a scored pose). Surface metrics score reference points visible from the reference camera path; precision, recall, and F@10 use a 10\,cm threshold~\cite{knapitsch2017tanks}. Maps are exported at 2\,cm on ORI and 4\,cm on EuRoC, with ScaRF's EuRoC clouds voxelized to match; the released ScaRF chunk evaluator uses 10\,m chunks and a 3\,cm threshold.
\begin{table}[h]
\centering
\caption{Reconstruction accuracy on ORI (medians). \emph{Global}: 2\,cm export, rigid trajectory alignment; mean distance (m) and F-score at 10\,cm. \emph{10\,m chunks}:
ScaRF evaluator, per-chunk $\Sim$+ICP; mean error (m), precision and recall (\%) at 3\,cm. \emph{Poses}: VIO = DAVIO's OpenVINS trajectory, Own = own trajectory, GT =
reference. Top: comparable methods (DA3-Base); \textbf{best}, \underline{second best}.
Bottom: not directly comparable, namely DAVIO with DA3-Large and the offline DAVIO and
ScaRF mappers with GT poses (\textbf{bold}: better of the two). DAVIO rows shaded.
$^\dagger$Global $\Sim$ fit to the reference.}
\label{tab:ori}
\scriptsize
\setlength{\tabcolsep}{4.5pt}
\newcommand{\grp}[1]{\multicolumn{7}{@{}l}{\textit{#1}}\\[1pt]}
\newcommand{\ours}{\rowcolor{gray!12}}
\begin{tabular}{@{}l c cc ccc@{}}
\toprule
 & & \multicolumn{2}{c}{Global} & \multicolumn{3}{c}{10\,m chunks} \\
\cmidrule(lr){3-4}\cmidrule(l){5-7}
Method & Poses & Mean$\downarrow$ & F@10$\uparrow$
       & Err.$\downarrow$ & P@3$\uparrow$ & R@3$\uparrow$ \\
\midrule
\grp{Comparable: estimated poses, rerun (conditioned maps use DA3-Base)}
Unconditioned reference       & VIO & 0.219 & 0.38 & 0.157 & 20.1 & 13.9 \\
\ours DAVIO online            & VIO & \textbf{0.080} & \textbf{0.75} & \underline{0.044} & \underline{60.8} & 38.8 \\
ScaRF mapper (offline)        & VIO & \underline{0.084} & \underline{0.69} & \textbf{0.043} & \textbf{61.6} & \underline{43.9} \\
MASt3R-Fusion                 & Own & 0.194 & 0.67 & 0.123 & 49.6 & \textbf{61.4} \\
VGGT-SLAM2$^\dagger$          & Own & 0.161 & 0.50 & 0.110 & 34.0 & 14.3 \\
\midrule
\grp{Not directly comparable}
\ours DAVIO online, DA3-Large & VIO & 0.061 & 0.81 & 0.033 & 71.9 & 38.3 \\
\addlinespace[3pt]
\ours DAVIO mapper (offline)  & GT  & 0.069 & 0.88 & 0.041 & 67.8 & \textbf{54.7} \\
ScaRF mapper (offline)        & GT  & \textbf{0.040} & \textbf{0.89} & \textbf{0.029} & \textbf{70.6} & 47.7 \\
\bottomrule
\end{tabular}
\end{table}


\section{Results}\label{sec:results}
\subsection{Q1: Earlier metric start-up}\label{sec:res_init}
\begin{table}[t]\centering\footnotesize
\caption{Start-up on 132 ten-second EuRoC windows. Success: published trajectory with ATE below 0.5\,m. $T_f$, $T_q$: median first-state and quality-penalized times. Grav., Vel., ATE: median gravity-direction, velocity, and trajectory errors of the published states.}
\scriptsize
\setlength{\tabcolsep}{2.5pt}
\begin{tabular}{llcccccc}\toprule
 & Initializer & Success & $T_f$ (s) & $T_q$ (s) & Grav.\ ($^\circ$) & Vel.\ (m/s) & ATE (m)\\\midrule
(a) & OpenVINS & 111/132 & 3.65 & 3.87 & 0.85 & 0.07 & 0.042\\
 & DAVIO & 109/132 & 2.43 & 2.66 & 0.99 & 0.07 & 0.044\\
\bottomrule\end{tabular}\\[3pt]
\begin{tabular}{llcccc}\toprule
 & & \multicolumn{2}{c}{Machine Hall} & \multicolumn{2}{c}{Vicon rooms}
 \\\cmidrule(lr){3-4}\cmidrule(lr){5-6}
(b) & Initializer & Success & $T_q$ (s) & Success & $T_q$ (s)\\\midrule
 & OpenVINS & 54/67 & 4.68 & 57/65 & 3.15\\
 & DAVIO & 56/67 & 2.30 & 53/65 & 4.50\\
\bottomrule\end{tabular}
\label{tab:init}
\end{table}
In Table~\ref{tab:init}a, DAVIO publishes its first state at a median of 2.43\,s against 3.65\,s and lowers the median $T_q$ from 3.87 to 2.66\,s. The gain is largest where the OpenVINS initializer must wait for parallax: in the Machine Hall (Table~\ref{tab:init}b), DAVIO halves $T_q$ (2.30 against 4.68\,s), qualifies 56/67 windows against 54/67, and is faster on every sequence by 0.9--2.9\,s. The overall success rate stays comparable (109/132 against 111/132.

\subsection{Q2: Causal and final trajectories}\label{sec:res_traj}
\begin{table}[t]\centering\footnotesize
\caption{EuRoC ATE (m). Native: OpenVINS~\cite{geneva2020openvins} alone, median of three mapping-disabled runs. Causal and final: map-frame output and final trajectory of one DAVIO run. Cov., $T_f$: DAVIO's sequence coverage and first-state time.}
\setlength{\tabcolsep}{3.5pt}\scriptsize
\begin{tabular}{lccccc}\toprule
 & \multicolumn{3}{c}{ATE (m)} & & \\\cmidrule(lr){2-4}
Seq. & Native & Causal & Final & Cov. & $T_f$ (s)\\\midrule
MH\_01 & 0.099 & 0.119 & 0.065 & 0.99 & 2.2\\
MH\_02 & 0.092 & 0.098 & 0.062 & 0.99 & 2.3\\
MH\_03 & 0.158 & 0.143 & 0.101 & 0.98 & 2.5\\
MH\_04 & 0.278 & 0.212 & 0.153 & 0.97 & 2.8\\
MH\_05 & 0.281 & 0.254 & 0.200 & 0.97 & 3.9\\
\bottomrule\end{tabular}

\label{tab:motion}
\end{table}
\begin{table}[t]\centering\scriptsize
\caption{End-to-end comparison on the EuRoC Machine Hall: ATE (m), rigid SE(3) alignment. OpenVINS: native trajectory (Table~\ref{tab:motion}); DAVIO: final trajectory; mono-in.: monocular--inertial, 74--82\,\% of the frames scored (largest map); rt: real-time output; kf: keyframes. $^\dagger$Global Sim(3) fit to the reference.}\label{tab:cmp_euroc_traj}
\setlength{\tabcolsep}{1.9pt}
\begin{tabular}{lcccccc}\toprule
 & OpenVINS & DAVIO & ORB-SLAM3 & \multicolumn{2}{c}{MASt3R-Fusion} & VGGT-SLAM2$^\dagger$\\\cmidrule(lr){5-6}
Seq. & native & final & mono-in. & rt & kf & kf\\\midrule
MH\_01 & 0.099 & \underline{0.065} & \textbf{0.058} & 0.419 & 0.344 & 0.188\\
MH\_03 & 0.158 & \underline{0.101} & \textbf{0.053} & 0.447 & 0.424 & 0.337\\
MH\_05 & 0.281 & \underline{0.200} & \textbf{0.060} & 0.555 & 0.513 & 0.989\\
\bottomrule
\end{tabular}
\end{table}

All eleven DAVIO runs complete (Table~\ref{tab:motion}), and DAVIO's initializer publishes first on MH\_01--04 (2.2--2.8\,s). The final trajectory lowers ATE by 29--45\,\% relative to the native trajectory on all five Machine Hall sequences.

\textbf{End-to-end comparison} (Table~\ref{tab:cmp_euroc_traj}). DAVIO publishes a pose for 97--99\,\% of each sequence, whereas ORB-SLAM3 scores 74--82\,\% of the frames. Against the learned dense systems, the native OpenVINS trajectory on which DAVIO builds has 2.0--4.2 times lower ATE than MASt3R-Fusion's real-time output, and DAVIO's final trajectory has 2.6--5.3 and 2.9--4.9 times lower ATE than the keyframes of MASt3R-Fusion and scale-corrected VGGT-SLAM2.

\subsection{Q3: Dense reconstruction}\label{sec:res_map}
\begin{figure*}[t]\centering
\includegraphics[width=0.95\textwidth]{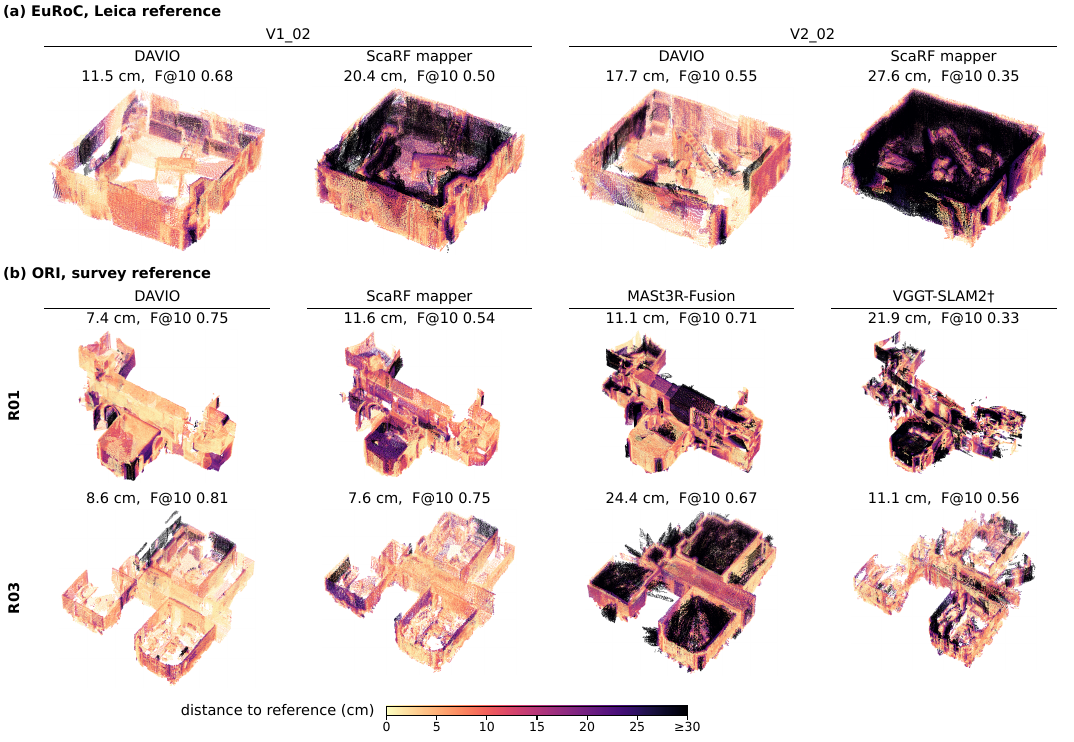}
\caption{Map-to-reference distance on (a)~EuRoC V1\_02 and V2\_02 (Leica scans) and (b)~ORI R01 and R03 (survey reference), with rigid trajectory alignment and no surface ICP. DAVIO and ScaRF's released mapper use the same OpenVINS trajectory and DA3-Base (matched poses); MASt3R-Fusion and VGGT-SLAM2 use their own trajectories, and VGGT-SLAM2 ($^\dagger$) is scaled by a global $\Sim$ fit to the reference. Titles: mean distance and F@10.}
\label{fig:errmap}
\end{figure*}
\begin{table}[t]
\centering\footnotesize
\caption{Map accuracy in the EuRoC Vicon rooms, evaluated against the Leica scans with
matched poses. DAVIO online and ScaRF's released mapper (offline) use the same OpenVINS
trajectory, DA3-Base, and 4\,cm density. F@10: F-score at 10\,cm; Mean: mean distance in metres. Best per sequence in
bold.}
\label{tab:cmp_euroc_map}
\setlength{\tabcolsep}{5pt}
\begin{tabular}{@{}l cc c cc@{}}
\toprule
 & \multicolumn{2}{c}{F@10 $\uparrow$} & & \multicolumn{2}{c}{Mean [m] $\downarrow$} \\
\cmidrule(lr){2-3}\cmidrule(lr){5-6}
Seq. & DAVIO & ScaRF & & DAVIO & ScaRF \\
\midrule
V1\_01 & \textbf{0.64} & 0.42          & & \textbf{0.125} & 0.242 \\
V1\_02 & \textbf{0.68} & 0.50          & & \textbf{0.115} & 0.204 \\
V1\_03 & \textbf{0.52} & 0.41          & & \textbf{0.205} & 0.288 \\
V2\_01 & \textbf{0.61} & 0.48          & & \textbf{0.167} & 0.210 \\
V2\_02 & \textbf{0.55} & 0.35          & & \textbf{0.177} & 0.276 \\
V2\_03 & 0.48          & \textbf{0.49} & & \textbf{0.188} & 0.211 \\
\midrule
Median & \textbf{0.58} & 0.45          & & \textbf{0.172} & 0.227 \\
\bottomrule
\end{tabular}
\end{table}
\textbf{EuRoC} (Table~\ref{tab:cmp_euroc_map}). On identical poses, DA3-Base checkpoint, and density, DAVIO online has a lower mean error than the released ScaRF mapper (offline) on all six Vicon-room sequences and a higher F@10 on five (on V2\_03, the F-scores differ by 0.01). The median F@10 rises from 0.45 to 0.58 and the median error falls from 0.227 to 0.172\,m; Fig.~\ref{fig:errmap}(a) shows the gap on V1\_02 and V2\_02. This comparison isolates the mapping layer.

\textbf{ORI} (Table~\ref{tab:ori}). On DAVIO's trajectory, pose conditioning reduces the median global error from 0.219 to 0.080\,m, with a lower error on every sequence (Fig.~\ref{fig:teaser}), and raises F@10 from 0.38 to 0.75 and chunk precision from 20.1 to 60.8\,\%. Running online, DAVIO outperforms ScaRF's offline mapper on the
same trajectory in global error and F@10 (0.080 vs.\ 0.084\,m; 0.75 vs.\ 0.69) at a comparable chunk error (0.044 vs.\ 0.043\,m), while ScaRF attains a higher recall. The end-to-end systems are less accurate: MASt3R-Fusion has 2.4 and 2.8 times DAVIO's global and chunk errors, and VGGT-SLAM2 reaches an F@10 of only 0.50 despite a global
$\Sim$ fit (Fig.~\ref{fig:errmap}b).

With reference poses, ScaRF's mapper is more accurate than DAVIO's offline mapper (0.040 vs.\ 0.069\,m global; 0.029 vs.\ 0.041\,m chunk), whereas DAVIO retains the higher recall. Moving both offline mappers from reference to OpenVINS poses, however, increases DAVIO's global error by 26\,\% (0.069 to 0.080\,m) but ScaRF's by 110\,\% (0.040 to 0.084\,m). DAVIO's advantage on estimated trajectories thus stems from its tolerance to
odometry drift, to which the submap graph contributes: with the graph poses fixed, the error on OpenVINS poses rises to 0.105\,m. Finally, DA3-Large lowers DAVIO's online errors to 0.061 and 0.033\,m and raises F@10 to 0.81; in the offline arm on identical fixed OpenVINS poses, it lowers the global error from 0.105 to 0.093\,m and raises chunk precision from 62.3 to 73.0\,\%. Depth quality is therefore a major remaining error source, alongside pose accuracy.
\subsection{Ablations}
\begin{table}[t]
\centering
\caption{Mapping ablations on EuRoC. F@10: F-score at 10\,cm; Compl.: median
completeness distance (mm); ATE: final absolute trajectory error on MH\_03 (m).
Rows marked $^\dagger$ replay the cached observations of the complete system with the
graph altered. Best per column in \textbf{bold}; the complete system is shaded.}
\label{tab:mapping}
\scriptsize
\setlength{\tabcolsep}{2.5pt}
\newcommand{\ours}{\rowcolor{gray!12}}
\begin{tabular}{@{}l ccc ccc c@{}}
\toprule
 & \multicolumn{3}{c}{F@10$\uparrow$} & \multicolumn{3}{c}{Compl.\,[mm]$\downarrow$}
 & ATE\,[m]$\downarrow$ \\
\cmidrule(lr){2-4}\cmidrule(lr){5-7}\cmidrule(l){8-8}
Variant & V1\_02 & V1\_03 & V2\_02 & V1\_02 & V1\_03 & V2\_02 & MH\_03 \\
\midrule
Unconditioned reference     & 0.514 & 0.432 & 0.493 & 42 & 67 & 38 & 0.158 \\
\ours DAVIO (complete)      & \textbf{0.685} & 0.520 & 0.550 & \textbf{30} & 44 & 31 & \textbf{0.101} \\
\addlinespace[2pt]
\quad w/o pose conditioning & 0.539 & 0.423 & 0.469 & 41 & 65 & 37 & 0.139 \\
\quad graph poses fixed$^\dagger$ & 0.682 & 0.519 & 0.547 & \textbf{30} & 44 & 31 & 0.158 \\
\quad w/o revisits$^\dagger$ & 0.575 & \textbf{0.529} & \textbf{0.599} & 31 & \textbf{40} & \textbf{28} & 0.158 \\
\bottomrule
\end{tabular}
\end{table}

Check Table~\ref{tab:mapping}. Pose conditioning is the main
source of surface quality. The complete system improves F@10 over the unconditioned
reference by 0.06--0.17 and completeness by 7--23\,mm. Removing conditioning alone
returns F@10 to within 0.03 of the reference. The graph, in contrast, mainly affects the
trajectory. Fixing its poses changes F@10 by at most 0.003, but it raises the final
MH\_03 ATE from 0.101 to 0.158\,m, the value of the uncorrected odometry. Removing
revisits yields the same 0.158\,m, so revisits provide the entire correction. Their
effect on surfaces depends on the sequence. They add 0.11 F@10 on V1\_02, but removing
them slightly improves V1\_03 and V2\_02 (by 0.009 and 0.049), possibly because some
admitted revisits impose imprecise scale constraints. On MH\_03, removing the drift gate
raises the final ATE to 0.116\,m.

\section{Conclusion}
DAVIO uses one feed-forward geometry model to start a classical VIO filter early and to build dense metric maps conditioned on its poses. It shortens Machine Hall start-up, lowers the final trajectory error, maps with lower error than ScaRF-SLAM's released mapper on identical EuRoC poses, and outperforms MASt3R-Fusion and VGGT-SLAM2 in EuRoC trajectory accuracy and ORI surface error.

\bibliographystyle{IEEEtran}
\bibliography{ref}

@IEEEtranBSTCTL{IEEEexample:BSTcontrol, CTLuse_forced_etal = "yes", CTLmax_names_forced_etal = "6", CTLnames_show_etal = "3"}

@article{martinelli2014closed,
  author   = {Martinelli, Agostino},
  title    = {Closed-form solution of visual-inertial structure from motion},
  journal  = {International Journal of Computer Vision},
  volume   = {106},
  number   = {2},
  pages    = {138--152},
  year     = {2014},
  doi      = {10.1007/s11263-013-0647-7}
}

@INPROCEEDINGS{wang2024dust3r,
  author={Wang, Shuzhe and Leroy, Vincent and Cabon, Yohann and Chidlovskii, Boris and Revaud, Jerome},
  booktitle={2024 IEEE/CVF Conference on Computer Vision and Pattern Recognition (CVPR)}, 
  title={DUSt3R: Geometric 3D Vision Made Easy}, 
  year={2024},
  volume={},
  number={},
  pages={20697-20709},
  doi={10.1109/CVPR52733.2024.01956}}

@InProceedings{leroy2024mast3r,
author="Leroy, Vincent
and Cabon, Yohann
and Revaud, Jerome",
editor="Leonardis, Ale{\v{s}}
and Ricci, Elisa
and Roth, Stefan
and Russakovsky, Olga
and Sattler, Torsten
and Varol, G{\"u}l",
title="Grounding Image Matching in 3D with MASt3R",
booktitle="Computer Vision -- ECCV 2024",
year="2025",
publisher="Springer Nature Switzerland",
address="Cham",
pages="71--91",
isbn="978-3-031-73220-1"
}

@INPROCEEDINGS{wang2025vggt,
  author={Wang, Jianyuan and Chen, Minghao and Karaev, Nikita and Vedaldi, Andrea and Rupprecht, Christian and Novotny, David},
  booktitle={2025 IEEE/CVF Conference on Computer Vision and Pattern Recognition (CVPR)}, 
  title={VGGT: Visual Geometry Grounded Transformer}, 
  year={2025},
  volume={},
  number={},
  pages={5294-5306},
  doi={10.1109/CVPR52734.2025.00499}}

@inproceedings{keetha2026mapanything,
  title={{MapAnything}: Universal Feed-Forward Metric {3D} Reconstruction},
  author={Nikhil Keetha and Norman M\"{u}ller and Johannes Sch\"{o}nberger and Lorenzo Porzi and Yuchen Zhang and Tobias Fischer and Arno Knapitsch and Duncan Zauss and Ethan Weber and Nelson Antunes and Jonathon Luiten and Manuel Lopez-Antequera and Samuel Rota Bul\`{o} and Christian Richardt and Deva Ramanan and Sebastian Scherer and Peter Kontschieder},
  booktitle={International Conference on 3D Vision (3DV)},
  year={2026},
  organization={IEEE}
}

@INPROCEEDINGS{he2023drt,
  author={He, Yijia and Xu, Bo and Ouyang, Zhanpeng and Li, Hongdong},
  booktitle={2023 IEEE/CVF Conference on Computer Vision and Pattern Recognition (CVPR)}, 
  title={A Rotation-Translation-Decoupled Solution for Robust and Efficient Visual-Inertial Initialization}, 
  year={2023},
  volume={},
  number={},
  pages={739-748},
  doi={10.1109/CVPR52729.2023.00078}}

@InProceedings{zhou2022learneddepthinit,
author="Zhou, Yunwen
and Kar, Abhishek
and Turner, Eric
and Kowdle, Adarsh
and Guo, Chao X.
and DuToit, Ryan C.
and Tsotsos, Konstantine",
editor="Avidan, Shai
and Brostow, Gabriel
and Ciss{\'e}, Moustapha
and Farinella, Giovanni Maria
and Hassner, Tal",
title="Learned Monocular Depth Priors in Visual-Inertial Initialization",
booktitle="Computer Vision -- ECCV 2022",
year="2022",
publisher="Springer Nature Switzerland",
address="Cham",
pages="552--570",
isbn="978-3-031-20047-2"
}

@ARTICLE{forster2017preintegration,
  author={Forster, Christian and Carlone, Luca and Dellaert, Frank and Scaramuzza, Davide},
  journal={IEEE Transactions on Robotics}, 
  title={On-Manifold Preintegration for Real-Time Visual--Inertial Odometry}, 
  year={2017},
  volume={33},
  number={1},
  pages={1-21},
  doi={10.1109/TRO.2016.2597321}}

@inproceedings{murai2025mast3rslam,
  title={{MASt3R-SLAM}: Real-Time Dense {SLAM} with {3D} Reconstruction Priors},
  author={Murai, Riku and Dexheimer, Eric and Davison, Andrew J.},
  booktitle={Proceedings of the IEEE/CVF Conference on Computer Vision and Pattern Recognition},
  year={2025},
}

@article{xin2023simplemapping,
      author    = {Xin, Yingye and Zuo, Xingxing and Lu, Dongyue and Leutenegger, Stefan},
      title     = {{SimpleMapping: Real-Time Visual-Inertial Dense Mapping with Deep Multi-View Stereo}},
      booktitle = {IEEE International Symposium on Mixed and Augmented Reality (ISMAR)},
      month     = {Oct},
      year      = {2023}
  }

@INPROCEEDINGS{rosinol2020kimera,
  author={Rosinol, Antoni and Abate, Marcus and Chang, Yun and Carlone, Luca},
  booktitle={2020 IEEE International Conference on Robotics and Automation (ICRA)}, 
  title={Kimera: an Open-Source Library for Real-Time Metric-Semantic Localization and Mapping}, 
  year={2020},
  volume={},
  number={},
  pages={1689-1696},
  doi={10.1109/ICRA40945.2020.9196885}}

@article{burri2016euroc,
  author  = {Burri, Michael and Nikolic, Janosch and Gohl, Pascal and Schneider, Thomas and Rehder, Joern and Omari, Sammy and Achtelik, Markus W and Siegwart, Roland},
  title   = {The EuRoC micro aerial vehicle datasets},
  year    = {2016},
  doi     = {10.1177/0278364915620033},
  journal = {The International Journal of Robotics Research}
}

@INPROCEEDINGS{zhang2018tutorial,
  author={Zhang, Zichao and Scaramuzza, Davide},
  booktitle={2018 IEEE/RSJ International Conference on Intelligent Robots and Systems (IROS)}, 
  title={A Tutorial on Quantitative Trajectory Evaluation for Visual(-Inertial) Odometry}, 
  year={2018},
  volume={},
  number={},
  pages={7244-7251},
  doi={10.1109/IROS.2018.8593941}}

@article{knapitsch2017tanks,
author = {Knapitsch, Arno and Park, Jaesik and Zhou, Qian-Yi and Koltun, Vladlen},
title = {Tanks and temples: benchmarking large-scale scene reconstruction},
year = {2017},
issue_date = {August 2017},
publisher = {Association for Computing Machinery},
address = {New York, NY, USA},
volume = {36},
number = {4},
issn = {0730-0301},
url = {https://doi.org/10.1145/3072959.3073599},
doi = {10.1145/3072959.3073599},
journal = {ACM Trans. Graph.},
month = jul,
articleno = {78},
numpages = {13},
}

@INPROCEEDINGS{potje2024xfeat,
  author={Potje, Guilherme and Cadar, Felipe and Araujo, André and Martins, Renato and Nascimento, Erickson R.},
  booktitle={2024 IEEE/CVF Conference on Computer Vision and Pattern Recognition (CVPR)}, 
  title={XFeat: Accelerated Features for Lightweight Image Matching}, 
  year={2024},
  volume={},
  number={},
  pages={2682-2691},
  doi={10.1109/CVPR52733.2024.00259}}

@InProceedings{lindenberger2023lightglue,
    author    = {Lindenberger, Philipp and Sarlin, Paul-Edouard and Pollefeys, Marc},
    title     = {LightGlue: Local Feature Matching at Light Speed},
    booktitle = {Proceedings of the IEEE/CVF International Conference on Computer Vision (ICCV)},
    month     = {October},
    year      = {2023},
    pages     = {17627-17638}
}

@article{lepetit2009epnp,
  author   = {Lepetit, Vincent and Moreno-Noguer, Francesc and Fua, Pascal},
  title    = {{EPnP}: An accurate {O(n)} solution to the {PnP} problem},
  journal  = {International Journal of Computer Vision},
  volume   = {81},
  number   = {2},
  pages    = {155--166},
  year     = {2009},
  doi      = {10.1007/s11263-008-0152-6}
}

@INPROCEEDINGS{mourikis2007msckf,
  author={Mourikis, Anastasios I. and Roumeliotis, Stergios I.},
  booktitle={Proceedings 2007 IEEE International Conference on Robotics and Automation}, 
  title={A Multi-State Constraint Kalman Filter for Vision-aided Inertial Navigation}, 
  year={2007},
  volume={},
  number={},
  pages={3565-3572},
  doi={10.1109/ROBOT.2007.364024}}

@ARTICLE{qin2018vinsmono,
  author={Qin, Tong and Li, Peiliang and Shen, Shaojie},
  journal={IEEE Transactions on Robotics}, 
  title={VINS-Mono: A Robust and Versatile Monocular Visual-Inertial State Estimator}, 
  year={2018},
  volume={34},
  number={4},
  pages={1004-1020},
  doi={10.1109/TRO.2018.2853729}}

@INPROCEEDINGS{geneva2020openvins,
  author={Geneva, Patrick and Eckenhoff, Kevin and Lee, Woosik and Yang, Yulin and Huang, Guoquan},
  booktitle={2020 IEEE International Conference on Robotics and Automation (ICRA)}, 
  title={OpenVINS: A Research Platform for Visual-Inertial Estimation}, 
  year={2020},
  volume={},
  number={},
  pages={4666-4672},
  doi={10.1109/ICRA40945.2020.9196524}}

@ARTICLE{campos2021orbslam3,
  author={Campos, Carlos and Elvira, Richard and Rodríguez, Juan J. Gómez and M. Montiel, José M. and D. Tardós, Juan},
  journal={IEEE Transactions on Robotics}, 
  title={ORB-SLAM3: An Accurate Open-Source Library for Visual, Visual–Inertial, and Multimap SLAM}, 
  year={2021},
  volume={37},
  number={6},
  pages={1874-1890},
  doi={10.1109/TRO.2021.3075644}}

@INPROCEEDINGS{peng2024sqrtvins,
  author={Peng, Yuxiang and Chen, Chuchu and Huang, Guoquan},
  booktitle={2024 IEEE International Conference on Robotics and Automation (ICRA)}, 
  title={Ultrafast Square-Root Filter-based VINS}, 
  year={2024},
  volume={},
  number={},
  pages={6966-6972},
  doi={10.1109/ICRA57147.2024.10610916}}

@INPROCEEDINGS{campos2020inertialonly,
  author={Campos, Carlos and Montiel, José M.M. and Tardós, Juan D.},
  booktitle={2020 IEEE International Conference on Robotics and Automation (ICRA)}, 
  title={Inertial-Only Optimization for Visual-Inertial Initialization}, 
  year={2020},
  volume={},
  number={},
  pages={51-57},
  doi={10.1109/ICRA40945.2020.9197334}}

@ARTICLE{yang2023selfcal,
  author={Yang, Yulin and Geneva, Patrick and Zuo, Xingxing and Huang, Guoquan},
  journal={IEEE Transactions on Robotics}, 
  title={Online Self-Calibration for Visual-Inertial Navigation: Models, Analysis, and Degeneracy}, 
  year={2023},
  volume={39},
  number={5},
  pages={3479-3498},
  doi={10.1109/TRO.2023.3275878}}

@article{zhang2026scarfslam,
  title={{ScaRF-SLAM}: Scale-Consistent Reconstruction with Feed-Forward Models and Classical Visual {SLAM}},
  author={Zhang, Yuhao and Tao, Yifu and Dellaert, Frank and Fallon, Maurice},
  journal={arXiv preprint arXiv:2606.00307},
  year={2026}
}

@article{maggio2026vggtslam2,
  title={{VGGT-SLAM 2.0}: Real-time Dense Feed-forward Scene Reconstruction},
  author={Maggio, Dominic and Carlone, Luca},
  journal={Robotics: Science and Systems},
  year={2026}
}

@article{maggio2025vggtslam,
  title={{VGGT-SLAM}: Dense RGB SLAM Optimized on the SL (4) Manifold},
  author={Maggio, Dominic and Lim, Hyungtae and Carlone, Luca},
  journal={Advances in Neural Information Processing Systems},
  volume={39},
  year={2025}
}

@article{wang2025pi3,
  title={PI3: Permutation-Equivariant Visual Geometry Learning},
  author={Wang, Yifan and Zhou, Jianjun and Zhu, Haoyi and Chang, Wenzheng and Zhou, Yang and Li, Zizun and Chen, Junyi and Pang, Jiangmiao and Shen, Chunhua and He, Tong},
  journal={arXiv preprint arXiv:2507.13347},
  year={2025}
}

@article{lin2025da3,
  title={Depth Anything 3: recovering the visual space from any views},
  author={Haotong Lin and Sili Chen and Jun Hao Liew and Donny Y. Chen and Zhenyu Li and Guang Shi and Jiashi Feng and Bingyi Kang},
  journal={arXiv preprint arXiv:2511.10647},
  year={2025}
}

@misc{zhang2026ffvioinit,
      title={Efficient Feature-Free Initialization for Monocular Visual-Inertial Systems Using a Feed-Forward 3D Model}, 
      author={Yuantai Zhang and Jiaqi Yang and Huajian Zeng and Changhao Chen and Haoang Li and Liang Li and Dezhen Song and Xingxing Zuo},
      year={2026},
      eprint={2605.17327},
      archivePrefix={arXiv},
      primaryClass={cs.RO},
}

@misc{cerezo2025closedform,
      title={An Efficient Closed-Form Solution to Full Visual-Inertial State Initialization}, 
      author={Samuel Cerezo and Seong Hun Lee and Javier Civera},
      year={2026},
      eprint={2511.18910},
      archivePrefix={arXiv},
      primaryClass={cs.RO},
}

@misc{hu2026ec3rslam,
      title={EC3R-SLAM: Efficient and Consistent Monocular Dense SLAM with Feed-Forward 3D Reconstruction}, 
      author={Lingxiang Hu and Naima Ait Oufroukh and Fabien Bonardi and Raymond Ghandour},
      year={2025},
      eprint={2510.02080},
      archivePrefix={arXiv},
      primaryClass={cs.RO},
}

@misc{zhou2026mast3rfusion,
      title={MASt3R-Fusion: Integrating Feed-Forward Visual Model with IMU, GNSS for High-Functionality SLAM}, 
      author={Yuxuan Zhou and Xingxing Li and Shengyu Li and Zhuohao Yan and Chunxi Xia and Shaoquan Feng},
      year={2025},
      eprint={2509.20757},
      archivePrefix={arXiv},
      primaryClass={cs.RO},
}

@misc{salih2026vidar,
      title={VIDAR: Visual-Inertial Dense Alignment and Reconstruction via a Geometric Foundation Model}, 
      author={Diyari Mohammed Salih and Lingxiang Hu and Naima AitOufroukh-Mammar and Fabien Bonardi},
      year={2026},
      eprint={2607.17171},
      archivePrefix={arXiv},
      primaryClass={cs.RO},
}

@misc{lee2026unisim,
      title={UniSim-SLAM: Feed-Forward SLAM with Unified Sim(3) Optimization}, 
      author={Inha Lee and Dongjae Jeong and Junhee Lee and Kyungdon Joo},
      year={2026},
      eprint={2608.01706},
      archivePrefix={arXiv},
      primaryClass={cs.CV},
}

@article{teed2021droid,
  title={{DROID-SLAM: Deep Visual SLAM for Monocular, Stereo, and RGB-D Cameras}},
  author={Teed, Zachary and Deng, Jia},
  journal={Advances in neural information processing systems},
  year={2021}
}

@INPROCEEDINGS{calibmorkis,
  author={Dong-Si, Tue-Cuong and Mourikis, Anastasios I.},
  booktitle={2012 IEEE/RSJ International Conference on Intelligent Robots and Systems}, 
  title={Estimator initialization in vision-aided inertial navigation with unknown camera-IMU calibration}, 
  year={2012},
  volume={},
  number={},
  pages={1064-1071},
  doi={10.1109/IROS.2012.6386235}}
\end{document}